\documentclass[11pt]{article}

\usepackage[preprint]{acl}
\usepackage{makecell}
\usepackage{tcolorbox}
\tcbuselibrary{skins, breakable, listings}
\usepackage{enumitem}
\usepackage{pifont} % 调用字体包
\usepackage{listings}
\usepackage{subcaption}
\usepackage{times}
\usepackage{latexsym}
\usepackage{booktabs}   % 提供 \toprule, \midrule, \bottomrule, \cmidrule
\usepackage{multirow}   % 提供 \multirow
\usepackage{array}      % 增强表格对齐和底层支持
\usepackage[T1]{fontenc}
\usepackage[utf8]{inputenc}

\usepackage{microtype}

\usepackage{inconsolata}

\usepackage{graphicx}
\usepackage{amsmath}
\usepackage{amssymb}
\usepackage{listings}
\usepackage{xcolor}
\usepackage{pifont}

\usepackage{booktabs}
\usepackage{multirow}
\usepackage[table]{xcolor}
\usepackage{fontawesome5}

\definecolor{snowblue}{HTML}{6F8FB5}
\definecolor{fireorange}{HTML}{D46A2C}

\newcommand{\nosft}{\textcolor{snowblue}{\faSnowflake}}
\newcommand{\sft}{\textcolor{fireorange}{\faFire}}
\definecolor{sectiongray}{HTML}{F7F7F8}
\definecolor{rulegray}{HTML}{949491}
\definecolor{agentgray}{HTML}{FCFDFF}
\definecolor{closedbg}{HTML}{FFFDF6}
\definecolor{sftgreen}{HTML}{F7FCF8}
\definecolor{agentavg}{HTML}{DDE2E8}
\definecolor{closedavg}{HTML}{F2EBCF}
\definecolor{sftavg}{HTML}{DFF1E6}

\title{HyperTool: Beyond Step-Wise Tool Calls for Tool-Augmented Agents}

\author{
 \textbf{Yaxin Du*\textsuperscript{1,2}},
 \textbf{Yifan Zhou*\textsuperscript{1}},
 \textbf{Yujie Ge\textsuperscript{1}},
 \textbf{Jiajun Wang\textsuperscript{1}},
\\
 \textbf{Xianghe Pang\textsuperscript{1}},
 \textbf{Shuo Tang\textsuperscript{1}},
 \textbf{Tuney Zheng\textsuperscript{2}},
 \textbf{Bryan Dai \textsuperscript{2}},
\\
 \textbf{Jian Yang\textsuperscript{3}},
 \textbf{Siheng Chen\textsuperscript{1}},
\\
 \textsuperscript{1}Shanghai Jiao Tong University,
 \textsuperscript{2}IQuest Research,
 \textsuperscript{3}Beijing University of Aeronautics and Astronautics,
\\
 \small{
    \textbf{* Equal Contribution} ;\  
   \textbf{Correspondence:} \href{sihengc@sjtu.edu.cn}{sihengc@sjtu.edu.cn}
 }
}

\begin{document}
\maketitle

\begin{abstract}
Tool-augmented LLM agents commonly rely on step-wise atomic tool calls, where each invocation, observation, and value transfer is exposed in the main reasoning trace. This creates an \emph{execution-granularity mismatch}: locally deterministic tool workflows are unfolded into repeated model-visible decisions, consuming context and forcing the model to manage low-level dataflow in the trace. We introduce \textbf{HyperTool}, a unified executable MCP-style tool interface that changes the model-visible unit of tool execution. A model invokes HyperTool with a code block that can call existing tools through their original schemas, manipulate returned values, and pass intermediate results locally, folding deterministic tool subroutines into a single outer call. To train models to use this interface, we synthesize HyperTool-format trajectories from cross-tool compositional tasks and verify them in real MCP environments. On MCP-Universe, HyperTool improves average accuracy from 15.69\% to 35.29\% on Qwen3-32B and from 9.93\% to 33.33\% on Qwen3-8B, and surpass GPT-OSS and Kimi-k2.5 on average accuracy, showing that our HyperTool can substantially improve multi-step tool use.
\end{abstract}

\section{Introduction}
\label{sec:intro}

Agentic systems powered by large language models (LLMs)~\citep{vaswani2017attention,touvron2023llama,openai2023gpt4,yang2024qwen2,yang2025qwen3} are increasingly applied across diverse environments. At their core, these agents integrate LLMs with a large set of tools, where each tool follows the MCP-style interface~\citep{MCP}: natural-language descriptions, input schemas, and structured calling formats. To solve tasks, agents typically inspect relevant information and perform reasoning step by step, invoking tools sequentially until the task is completed~\citep{yao2022react,schick2023toolformer,qin2023toolllm,mialon2023augmented}. This interface is simple and general, but it makes every atomic tool call a model-visible transition: each observation is written back into the main trace before the next action.

\begin{figure}[t]
\centering 
\includegraphics[width=1\linewidth]{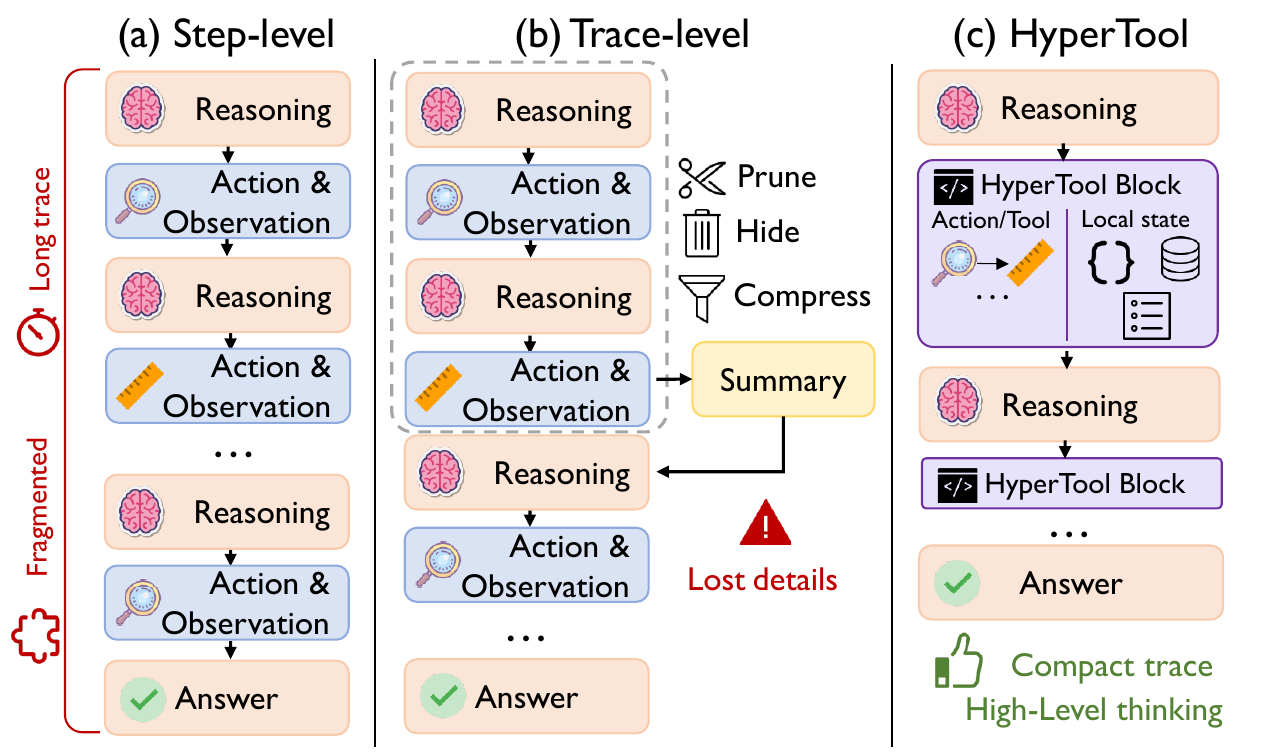} 
% \vspace{-0.8cm}
\caption{Comparison of context management paradigms in tool-augmented agents.
(a) Step-wise execution expands atomic calls and observations into the trace.
(b) Trace-level compression shortens the trace after expansion.
(c) HyperTool manages context at execution time by folding dependent tool operations into one executable block and returning only the task-relevant result.} 
\label{fig:placeholder} 
% \vspace{-0.5cm}
\end{figure}

% This coupling becomes inefficient when several tool calls form a locally deterministic subroutine. In such cases, step-wise invocation forces the agent to maintain each intermediate operation in the main reasoning trace, even when the intermediate results are only needed by subsequent tools rather than by the agent itself\cite{yao2022react}. For example, computing the distance between two addresses may require two geocoding calls and a distance calculation, while the main reasoning trace only needs the final distance. 

The atomic design of current tool interfaces conflicts with the compositional demand of real tool-use tasks. The issue is that a task-level operation is often implemented as a chain of lower-level tool calls, while the outer reasoning process only needs the composed result.
For example, answering the distance between two places may require geocoding both locations and then computing their distance, but the agent does not need to reason over the intermediate coordinates themselves.
In the step-wise format, this single distance-query operation is unfolded into multiple model-visible transitions, exposing geocoding outputs, coordinates, and the final calculation separately in the main trace. This creates two central challenges. First, it causes \emph{context inflation}: long tool observations accumulate in the reasoning trace even when only a few fields are actually needed\cite{qin2023toolllm}. Second, it causes \emph{reasoning fragmentation}: the model must repeatedly return to the main decision loop for low-level deterministic operations, interleaving high-level task reasoning with procedural state transfer\cite{wang2024executable}. Although these sub-tasks are ultimately executed by tools, explicitly unfolding their full call sequence in the main trace disrupts the overall reasoning structure and imposes unnecessary cognitive burden on the agent.

Prior work on tool-agent context management mostly operates at the \emph{trace level}. A direct approach is context pruning or compression, which reduces trace length but may discard intermediate values needed by later steps\cite{jiang2023llmlingua}. 
More structured methods use task-specific caching or summarization to retain compact state, but often rely on workflow-specific assumptions and generalize poorly across heterogeneous tools\cite{packer2023memgpt}. 
Subagent-based methods go further by hiding low-level execution from the main context, but this coarser delegation reduces transparency and makes intermediate control or recovery more difficult\cite{wu2024autogen,shen2023hugginggpt}. 
Overall, these methods reduce context overhead by discarding, compressing, or hiding execution traces, but they do not change the underlying execution granularity of tool use.

To move context management from post-hoc trace compression to execution-time control, we propose \textbf{HyperTool}, a unified MCP-compatible executable interface whose input is a code block. Inside the block, existing tools can be invoked through their original schemas, while intermediate outputs can be stored, transformed, filtered, and passed across calls using ordinary program logic. This unified format supports both atomic tool calls and multi-tool subroutines within the same interface. By shifting local execution into the block, HyperTool improves context management and tool-use expressivity at the same time: transient observations and deterministic state transfer stay out of the main trace, while code can compose tools, transform intermediate values, and define temporary helper functions. This allows the agent to maintain a compact high-level trace while constructing task-specific partial tools that are difficult to express through isolated atomic calls.

To make models learn this capability, we construct HyperTool-format trajectories for supervised fine-tuning. Exposing HyperTool only defines a new action space; the model must still learn when to invoke a block, how to compose tools inside it, and when to return intermediate observations to the main reasoning trace. We therefore synthesize cross-tool compositional tasks, roll out HyperTool trajectories in real MCP environments, and keep only execution-correct and evidence-consistent examples. This shifts supervision from isolated atomic calls to executable tool subroutines, teaching the model both block construction and appropriate tool-operation grouping.

We evaluate HyperTool on MCP-Universe using Qwen3-8B and Qwen3-32B. 
Finetuned on HyperTool trajectories substantially improves average accuracy, reaching 33.33\% on Qwen3-8B and 35.29\% on Qwen3-32B, compared with 9.93\% and 15.69\% for the corresponding base models. 
Notably, the Qwen3-32B HyperTool model is competitive with recent agentic/open models such as GPT-OSS and Kimi-k2.5, while outperforming all non-HyperTool baselines built on the same Qwen3 backbones. 

Our contributions are summarized as follows:

\begin{itemize}[topsep=0.1em,itemsep=0.01em,leftmargin=1em]
    \item We identify overly fine-grained tool execution as a key limitation: step-wise calls expose intermediate observations and tool dependencies to the model, inflating the context and fragmenting reasoning.
    \item We propose \textbf{HyperTool}, a unified executable tool interface that preserves the original MCP tool schema while allowing atomic calls and locally deterministic multi-tool workflows to be expressed within the same code-block format.
    \item We construct verified HyperTool-format trajectories for supervised fine-tuning and show consistent gains on MCP-Universe, improving average accuracy from 9.93\% to 33.33\% on Qwen3-8B.
\end{itemize}

% \vspace{-0.1cm}
\section{Related Work}
\label{sec:related}
% \vspace{-0.1cm}
\subsection{Tool-Augmented Agents and Tool Interfaces}
% \vspace{-0.1cm}

Tool-augmented agents extend LLMs by allowing models to interact with external tools, APIs, and environments during inference. Prior work has studied API-style tool use through reasoning-action interleaving, self-supervised tool-call learning, and instruction-tuned tool-use trajectories~\citep{yao2022react,schick2023toolformer,qin2023toolllm}. Recent tool ecosystems and benchmarks further scale this setting to large API and MCP-style tool spaces~\citep{li2023apibank,qin2023toolllm,basu2024nestful,luo2025mcpuniverse,mo2025livemcpbench,wang2025mcp,du2025infomosaic,wang2026mcppersonabenchmarkingllmagents}.  These works establish a common interface in which tools are described by schemas and invoked through structured calls. HyperTool is complementary rather than alternative to this interface: it is exposed as an MCP-style tool and invokes existing tools through their original schemas. Its goal is not to change how tools are specified, but to change the execution granularity of tool use, allowing local deterministic workflows to be executed inside one block instead of being exposed as repeated model-visible calls.

\subsection{Agent Execution Abstractions}

A second line of work abstracts agent execution to reduce the context cost of long, tool-intensive workflows.
These methods operate at different execution levels. 
\ding{172} Program-based agents use code as an expressive action representation, allowing models to combine computation, control flow, and environment interaction within executable programs~\citep{gao2023pal,chen2023program,wang2024executable,wang2023voyager,yu2025recode}. While powerful, they typically treat code as a general action language or task artifact, rather than as a wrapper for grouping existing tool calls. \ding{173} Workflow-level methods compile execution structures, such as dependency graphs or replayable workflow blueprints, before execution~\citep{kim2024llmcompiler,parmar2026separating,du2026swedevevaluatingtrainingautonomous,ye2025masgpttrainingllmsbuild,du2026g2readerdualevolvinggraphs}. 
They reduce repeated model invocation at the plan level, but assume the workflow can be specified upfront.
\ding{174} Trace-level context-management methods reduce long interaction histories through pruning, compression, summarization, memory modules, subagents, or context folding~\citep{sun2025scaling,ye2025agentfold,jiang2023llmlingua}. These methods manage the trace after step-wise execution has exposed intermediate observations. 
HyperTool abstracts a different level: it keeps MCP tools explicit, but changes the model-visible unit of tool execution by folding locally deterministic tool chains into one executable block.

\section{Method}
\label{sec:method}
\subsection{Problem Formulation}

\begin{figure*}[t]
    \centering
    \includegraphics[width=\textwidth]{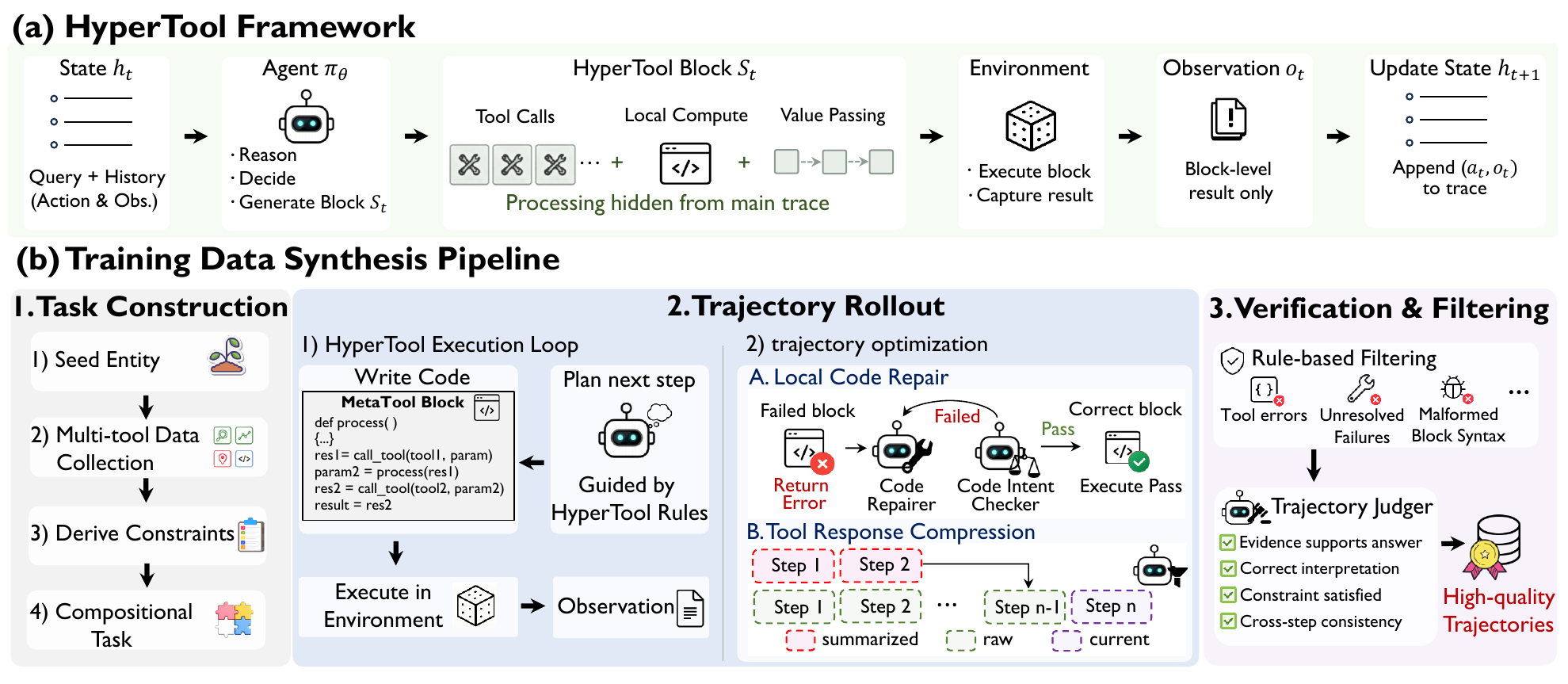}
    % \vspace{-0.9cm}
    \caption{Overview of the HyperTool framework and data construction pipeline. 
    (a) We synthesize compositional tasks, roll out HyperTool-format trajectories with local repair and trajectory-level context compression, and verify the resulting traces for execution correctness and evidence consistency.
    (b) HyperTool is exposed as an executable MCP-style tool. The agent writes a code block that invokes existing tools, performs local computation, and passes intermediate values internally. Only the final block-level observation is returned to the main reasoning trace.}
    \label{fig:overview}
    % \vspace{-0.3cm}
\end{figure*}

Tool-augmented agents solve tasks by repeatedly interacting with external tools.
At each step, the model chooses an action, the tool environment executes it, and the returned observation is appended to the execution trace.
We formulate this process as an interaction between a model and a tool environment.

Given a user query $q$ and a tool set $\mathcal{T}$ with specification $\Phi(\mathcal{T})$, the agent maintains an execution trace
$h_t = (q, a_{<t}, o_{<t})$, 
where $a_{<t}$ and $o_{<t}$ denote previous actions and observations.
At step $t$, the model emits an action

% \vspace{-0.3cm}
$$a_t \sim \pi_\theta(\cdot \mid h_t, \Phi(\mathcal{T})),$$
which is executed by the environment to produce an observation $o_t$.
The trace is then updated as
$h_{t+1} = h_t \cup \{(a_t, o_t)\}$.

Under the standard MCP-style interface, each action is an atomic tool call:

% \vspace{-0.3cm}
\begin{equation*}
a_t = \mathrm{Call}(\tau_t, x_t), \qquad \tau_t \in \mathcal{T},
\end{equation*}
where $\tau_t$ is the selected tool and $x_t$ is its input.
Thus, a workflow with $k$ dependent tool operations must be expressed as $k$ model-visible transitions.
This atomic execution granularity is simple and general, but it exposes every intermediate observation, value transfer, and procedural step to the main reasoning trace.

\subsection{HyperTool Framework}

HyperTool changes the model-visible unit of tool execution.
Instead of requiring every MCP-style tool operation to appear as a separate transition, HyperTool exposes a unified executable interface:

% \vspace{-0.3cm}
\begin{equation*}
a_t = \mathrm{Block}(S_t),
\end{equation*}
where $S_t$ is a bounded executable program written over the primitive tools in $\mathcal{T}$.
The HyperTool runtime executes $S_t$, allows existing tools to be invoked through their original schemas, and returns a single block-level observation $o_t$ to the outer trace.

Inside a HyperTool block, the model can call existing tools, store returned values as local variables, parse and filter observations, perform lightweight computation, define temporary helper functions, and pass intermediate values directly across calls.
This format subsumes ordinary atomic tool calling: if $S_t$ contains only one primitive call $\mathrm{Call}(\tau_t,x_t)$, HyperTool reduces to the standard tool-call interface.
If $S_t$ contains multiple dependent calls, HyperTool expresses them as one executable subroutine rather than multiple model-visible transitions.

The intended use case is a locally deterministic tool workflow whose intermediate observations support local dataflow but do not require task-level deliberation.
For example, a block may retrieve candidates, parse fields, filter results, compute derived values, and aggregate evidence before returning a compact result.
These intermediate states can remain inside the executable block.
In contrast, observations that may change the task plan, reveal an unknown schema, invalidate the current direction, or require semantic judgment should be returned to the main reasoning trace.
Thus, HyperTool does not replace the underlying MCP tools; it changes which parts of tool execution are exposed as separate model-visible transitions.

\subsection{Data Synthesis}
% \vspace{-0.1cm}

The HyperTool interface defines a new action format, but the model must still learn how to use it.
In particular, it must learn three behaviors: when to invoke a HyperTool block, how to implement internal tool composition inside the block, and when intermediate observations should return to the outer reasoning trace rather than remain local.

We construct HyperTool-format training trajectories through a three-stage pipeline:
\emph{compositional task construction}, \emph{HyperTool trajectory rollout}, and \emph{execution-and-evidence verification}.
The resulting data provides supervision at the level of executable tool subroutines rather than isolated atomic calls.
This teaches the model to internalize locally deterministic tool workflows while preserving task-level reasoning in the main trace.

\subsubsection{Compositional Task Construction}
% \vspace{-0.1cm}

The first stage constructs training questions that require genuine multi-tool composition.
For each domain, we start from real entities as seeds and let a teacher model interact with the corresponding MCP tool environment.
During this process, the model invokes domain-relevant tools to collect factual attributes from multiple sources and dimensions.

Based on the collected tool outputs, the model induces a set of mutually independent but complementary constraints.
These constraints are grounded in real returned values: numerical thresholds, entity attributes, identifiers, temporal information, and other condition values are derived from tool observations rather than hallucinated.
We require the constraints to span different tools or data dimensions, so that no single primitive call can directly determine the answer.
Instead, the answer must be identified through cross-source composition, local filtering, comparison, or aggregation.

This construction makes the intended solution path compositional.
A correct trajectory must coordinate multiple tools, pass intermediate values across calls, and perform secondary processing over returned results.
These tasks therefore create natural opportunities for HyperTool blocks, where dependent tool operations can be implemented as local executable subroutines.

\subsubsection{HyperTool Trajectory Collection}
% \vspace{-0.1cm}

Training HyperTool requires trajectories that go beyond ordinary stepwise tool calls, teaching the model not only which tools to call but also how to group dependent operations and where to place execution boundaries. 
We design a data collection pipeline that produces boundary-aware HyperTool demonstrations through three complementary mechanisms: \emph{context compression}, \emph{local repair}, and \emph{trajectory verification}.

\noindent \textbf{HyperTool Trajectory Rules}
We treat HyperTool as a directly callable MCP-style tool, while the underlying tools are provided in the system prompt with their names, input schemas, and expected outputs.
We use three rules to guide trajectory generation.
(1) \emph{Reasoning Precedence}: Before each tool call, the model generates reasoning content related to the current operation rather than invoking the tool directly.
(2) \emph{Blockwise Execution}: When a task can be decomposed into deterministic tool calls, the model should combine multiple calls and simple processing steps into a single HyperTool code block; if subsequent steps require complex semantic reasoning or outputs are uncertain, the model should call tools step by step.
(3) \emph{Internal Logic}: Each HyperTool block must be self-contained, contain only executable logic, and assign the final tool output or processed result to the variable `result`. Status-only strings or `print` outputs must not be used as final results.
These rules collectively constrain tool usage, reasoning generation, and output format, producing HyperTool trajectories that are structured, traceable, and suitable for training.

\noindent \textbf{Context Compression}
During rollout, long trajectories may contain many previous tool responses that are no longer central to the current decision.
To keep data generation focused on the current step, we maintain two traces: a full trace for archival and final data storage, and a denoised trace for subsequent model prompting.
Older tool responses in the denoised trace are summarized using a language model conditioned on the task, prior trajectory, tool name, tool arguments, and full tool output.
The summary retains information relevant for task solving, preserves key values such as numbers, URLs, and entity names, and removes format noise or irrelevant details. This compression is used only during trajectory synthesis.
Its role is to fold earlier rollout context so that the teacher model can focus on the current HyperTool decision while the full original trace remains available for verification and data storage.

\noindent \textbf{Local Repair of Failed Tool Blocks}
Generated HyperTool blocks may fail because of syntax errors, schema mismatches, or runtime errors. When a block fails, the system records the original assistant message, code, and error, and constructs a repair prompt for the model.
A static intent check ensures the repaired code still targets the same resource and operation type; if not, the model retries, with a bounded number of attempts to prevent infinite loops.
Successful repairs are executed and appended to the trace, while failures are preserved to avoid polluting prior or subsequent steps.

% \vspace{-0.1cm}
\subsubsection{Trajectory Verification}
% \vspace{-0.1cm}

Candidate trajectories are not used directly as supervision.
We verify them through execution-correctness filtering and evidence-consistency checking.

\noindent \textbf{Execution-correctness filtering.}
We remove trajectories with structural or runtime defects, including malformed block syntax, missing tool invocations, environment-level tool errors, unresolved execution failures, and other obvious anomalies.
This step ensures that retained trajectories demonstrate valid HyperTool execution rather than accidental code behavior.

\noindent \textbf{Evidence-consistency verification.}
We then verify the full trajectory with an LLM judge.
The judge checks whether the final answer is supported by returned tool evidence, whether intermediate results are interpreted correctly, whether conclusions remain consistent across blocks, and whether the final response satisfies all task constraints.
To reduce single-run variance, we query the judge multiple times and retain only trajectories that pass majority-vote verification.

\begin{table*}[t]
\centering
\small
\renewcommand{\arraystretch}{0.98}
\setlength{\tabcolsep}{4.2pt}
\caption{Main results on MCP-Universe. We compare two groups of systems: off-the-shelf agentic models used without task-specific fine-tuning, and Qwen3-based baselines. SFT denotes supervised fine-tuning; \nosft{} indicates models evaluated without SFT, while \sft{} indicates models fine-tuned on our collected HyperTool trajectory data. Within each Qwen3 backbone group, \textbf{bold} marks the best result and \underline{underline} marks the second-best result.}
% \vspace{-0.3cm}
\label{tab:main_results}
\resizebox{0.95\linewidth}{!}{%
\begin{tabular}{llccccccc}
\toprule
\multirow{2}{*}{\textbf{Model}} & \multirow{2}{*}{\textbf{Method}} & \multirow{2}{*}{\textbf{SFT}} & \multicolumn{6}{c}{\textbf{MCP-Universe}} \\
\cmidrule(lr){4-9}
& & & \makecell{\textbf{Financial}\\\textbf{Analysis}} & \makecell{\textbf{Repository}\\\textbf{Management}} & \makecell{\textbf{Location}\\\textbf{Navigation}} & \makecell{\textbf{Web}\\\textbf{Search}} & \makecell{\textbf{Avg.}\\\textbf{Acc.}} & \makecell{\textbf{Avg.}\\\textbf{Score}} \\
\arrayrulecolor{rulegray}
\specialrule{0.35pt}{0pt}{0pt}
\rowcolor{sectiongray}\multicolumn{9}{c}{\textit{Agentic Models}} \\
\specialrule{0.35pt}{0pt}{0pt}
\rowcolor{agentgray} GPT-OSS & -- & \nosft & 62.50 & 14.28 & 25.71 & 26.00 & \cellcolor{agentavg}32.13 & \cellcolor{agentavg}50.43 \\
\rowcolor{agentgray} GLM-5.1 & -- & \nosft & 55.00 & 42.86 & 28.57 & 36.00 & \cellcolor{agentavg}36.60 & \cellcolor{agentavg}53.40 \\
% \rowcolor{agentgray} GPT-5 & -- & \nosft & 62.50 & 42.86 & 28.57 & 56.00 & \cellcolor{agentavg}49.02 & \cellcolor{agentavg}68.97 \\
\rowcolor{agentgray} Gemini-2.5-Flash & -- & \nosft & 25.00 & 30.77 & 30.77 & 19.51 & \cellcolor{agentavg}25.58 & \cellcolor{agentavg}39.84 \\
\rowcolor{agentgray} Kimi-k2.5 & -- & \nosft & 51.28 & 44.44 & 17.14 & 25.00 & \cellcolor{agentavg}34.17 & \cellcolor{agentavg}56.96 \\
\rowcolor{agentgray} DeepSeek-V4-Flash & -- & \nosft & 57.50 & 39.29 & 29.41 & 45.83 & \cellcolor{agentavg}44.00 & \cellcolor{agentavg}47.22 \\
\specialrule{0.35pt}{0pt}{0pt}
\specialrule{0.35pt}{0pt}{0pt}
\rowcolor{sectiongray}\multicolumn{9}{c}{\textit{Baselines}} \\
\specialrule{0.35pt}{0pt}{0pt}
\rowcolor{closedbg} & Base & \nosft & 17.14 & 3.57 & 17.14 & 4.00 & \cellcolor{closedavg}9.93 & \cellcolor{closedavg}28.75 \\
\rowcolor{closedbg} & AgentFold~\citep{ye2025agentfold} & \nosft & 27.78 & 17.81 & \underline{25.71} & 16.00 & \cellcolor{closedavg}21.63 & \cellcolor{closedavg}37.81 \\
\rowcolor{closedbg} & ReCode~\citep{yu2025recode} & \nosft & 25.00 & 10.71 & 14.29 & 10.00 & \cellcolor{closedavg}15.03 & \cellcolor{closedavg}32.70 \\
\rowcolor{closedbg} & BrowseMaster~\citep{pang2025browsemaster} & \nosft & 21.50 & 14.29 & 11.43 & 4.00 & \cellcolor{closedavg}18.60 & \cellcolor{closedavg}30.97 \\
\rowcolor{closedbg} & CodeAct~\citep{wang2024executable} & \nosft & 20.00 & 14.29 & 22.86 & \textbf{22.00} & \cellcolor{closedavg}20.92 & \cellcolor{closedavg}38.75 \\
\rowcolor{closedbg} & ReAct~\citep{yao2022react} & \sft & 32.50 & 7.14 & \underline{25.71} & 16.00 & \cellcolor{closedavg}20.92 & \cellcolor{closedavg}35.09 \\
\rowcolor{closedbg}\cellcolor{closedbg}\multirow{-7}{*}{Qwen3-8B} & \textbf{HyperTool (Ours)} & \sft & \textbf{62.50} & \underline{25.00} & \underline{28.57} & 18.00 & \cellcolor{closedavg}\underline{33.33} & \cellcolor{closedavg}\underline{48.42} \\
\specialrule{0.35pt}{0pt}{0pt}
\rowcolor{sftgreen} & Base & \nosft & 35.00 & 3.57 & 20.00 & 4.00 & \cellcolor{sftavg}15.69 & \cellcolor{sftavg}31.99 \\
\rowcolor{sftgreen} & AgentFold~\citep{ye2025agentfold} & \nosft & \underline{52.40} & \textbf{26.94} & 14.30 & 16.00 & \cellcolor{sftavg}27.13 & \cellcolor{sftavg}42.20 \\
\rowcolor{sftgreen} & ReCode~\citep{yu2025recode} & \nosft & 40.00 & 25.00 & 11.43 & 6.00 & \cellcolor{sftavg}19.60 & \cellcolor{sftavg}38.41 \\
\rowcolor{sftgreen} & BrowseMaster~\citep{pang2025browsemaster} & \nosft & 37.50 & 21.43 & 17.14 & 10.00 & \cellcolor{sftavg}20.91 & \cellcolor{sftavg}36.16 \\
\rowcolor{sftgreen} & CodeAct~\citep{wang2024executable} & \nosft & 42.50 & \underline{25.00} & 14.29 & \underline{20.00} & \cellcolor{sftavg}25.49 & \cellcolor{sftavg}42.04 \\
\rowcolor{sftgreen} & ReAct~\citep{yao2022react} & \sft & 40.00 & 21.43 & 20.00 & 16.00 & \cellcolor{sftavg}24.18 & \cellcolor{sftavg}41.28 \\
\rowcolor{sftgreen}\cellcolor{sftgreen}\multirow{-7}{*}{Qwen3-32B} & \textbf{HyperTool (Ours)} & \sft & \textbf{62.50} & 21.43 & \textbf{34.29} & \textbf{22.00} & \cellcolor{sftavg}\textbf{35.29} & \cellcolor{sftavg}\textbf{53.18} \\
\arrayrulecolor{black}
\bottomrule
\end{tabular}%
}
% \vspace{-0.3cm}
\end{table*}

% \vspace{-0.1cm}
\subsection{Supervised Fine-Tuning}
% \vspace{-0.1cm}

After verification, we convert each retained trajectory into a supervised training example using the same chat/tool format as inference.
The loss is applied to model-generated assistant turns, including reasoning text and HyperTool code blocks.
Tool observations are treated as environment outputs and excluded from the prediction target.

Training on these trajectories teaches the model to emit executable HyperTool blocks, compose existing MCP tools through local program logic, and decide whether subsequent operations should remain inside the block or return to step-wise interaction.
This supervision shifts learning from predicting isolated atomic tool calls to learning executable tool subroutines for multi-step tool use.

% \vspace{-0.1cm}
\section{Experiments}
\label{sec:experiments}
% \vspace{-0.05cm}

\noindent\textbf{Implementation}. We train our HyperTool based on open-source LLM Qwen3~\citep{yang2025qwen3} series model with 8B parameters and 32B parameters in total respectively. We set the max tool call number as 50 and max context tokens as 128k, any trajectory exceed any of these settings will be forcibly terminated.

\noindent\textbf{Benchmark}. We compare our method with other baselines on a general benchmark: MCP-Universe~\citep{luo2025mcpuniverse}. We evaluate on four domains: web search, financial analysis, location navigation and repository management to test the general capabilities of agents.

\noindent\textbf{Baseline}. We compare HyperTool with representative agent frameworks and execution abstractions for multi-step reasoning and environment interaction. Specifically, we include ReAct~\citep{yao2022react}, which performs step-by-step atomic tool interaction; CodeAct~\citep{wang2024executable} and ReCode~\citep{yu2025recode}, which formulate agent behavior through executable code generation; BrowseMaster~\citep{pang2025browsemaster}, a programmatic web-browsing agent framework; and AgentFold~\citep{ye2025agentfold}, which improves long-horizon agent execution through proactive context management. These baselines represent different approaches to reducing the limitations of purely atomic execution and provide strong comparisons for evaluating HyperTool’s execution abstraction.

% \vspace{-0.05cm}
\subsection{Main Results.}
% \vspace{-0.05cm}

To evaluate whether HyperTool improves MCP-style multi-tool execution, we conduct all experiments on MCP-Universe under the same evaluation budget: each trajectory is limited to at most 50 tool calls and 128k context tokens. Table~\ref{tab:main_results} reports performance on four domains, together with average accuracy and average score. We compare two settings: off-the-shelf agentic models evaluated without task-specific fine-tuning, and Qwen3-based methods, including both non-SFT baselines and SFT models. For supervised training, we use GLM-5.1~\citep{zeng2026glm} to synthesize training tasks, generate trajectories, and filter the resulting data. To ensure a fair comparison, the ReAct-SFT baseline uses the same training tasks and the same GLM-5.1-based trajectory generation and filtering pipeline as HyperTool.

The results reveal four main findings. \ding{182} HyperTool outperforms all Qwen3-based execution and context-management baselines in the overall metrics on both 8B and 32B backbones, reaching 48.42 average score on 8B and 53.18 on 32B. \ding{183} For a fair SFT comparison, we also train a ReAct model using the same synthesized tasks, the same trajectory generator, and the same filtering strategy. HyperTool still substantially outperforms ReAct-SFT, improving average accuracy from 20.92\% to 33.33\% on 8B and from 24.18\% to 35.29\% on 32B. \ding{184} The largest gains appear in composition-heavy domains such as Financial Analysis, where both scales reach 62.50\% accuracy, suggesting that executable local blocks are useful for value passing, filtering, and aggregation across tool outputs. \ding{185} This superior performance is directly driven by HyperTool's expanded tool orchestration capacity; as detailed in Table~\ref{tab:tool_usage_stats} in the Appendix, it executes significantly more primitive tools within fewer interaction turns, effectively bypassing the context bottlenecks that typically truncate long ReAct trajectories. \ding{186} Even with the 8B backbone, HyperTool narrows the gap to strong off-the-shelf agents, surpassing GPT-OSS (32.13\%) and Gemini-2.5-Flash (25.58\%) in average accuracy.

\begin{figure}[t]
    \centering

    \begin{subfigure}[t]{0.485\columnwidth}
        \centering
        \includegraphics[width=1\linewidth]{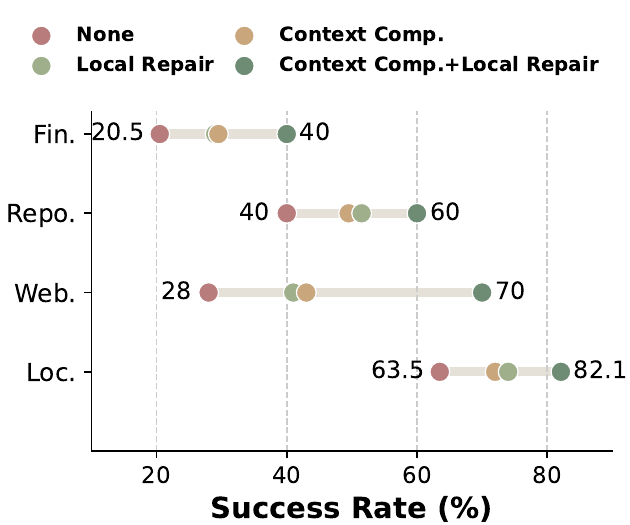}
        \caption{Data Synthesis}
        \label{fig:ablation_synthesis}
    \end{subfigure}%
    \hspace{0.01\columnwidth}%
    \begin{subfigure}[t]{0.505\columnwidth}
        \centering
        \includegraphics[width=1\linewidth]{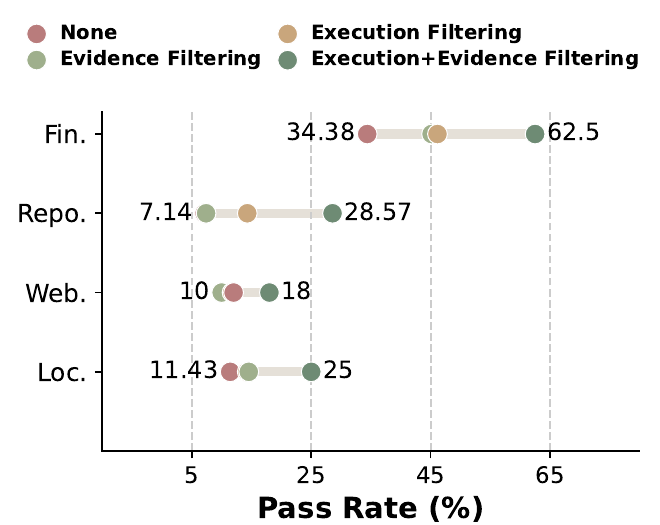}
        \caption{Trajectory Verification}
        \label{fig:ablation_verification}
    \end{subfigure}
    % \vspace{-0.3cm}
    \caption{Ablation study on the HyperTool components. (a) The impact of Context Compression and Local Repair on the data synthesis success rate. (b) The effect of Execution and Evidence Filtering on the sft model performance.}
    \label{fig:ablation_study}
    % \vspace{-0.2cm}
\end{figure}

% \vspace{-0.05cm}
\subsection{Ablation studies}
% \vspace{-0.05cm}

\noindent \textbf{Execution Interface Ablation}
To determine the optimal execution interface design, we conduct an ablation study on the Qwen3-8B backbone, evaluating whether HyperTool should serve as a unified interface for all interactions (\textit{HyperTool-only}) or be reserved exclusively for multi-step workflows (\textit{Atomic+HyperTool} hybrid). Table~\ref{tab:ablation_interface} presents the performance across these configurations, revealing two clear trends: 
\textbf{\ding{182}} \textbf{A unified action space maximizes performance.} The \textit{HyperTool-only} mode achieves a significantly higher average accuracy (33.33\%) compared to the hybrid setting (26.85\%) and the ReAct baseline (20.92\%). This indicates that forcing smaller models to dynamically classify and switch between different execution interfaces introduces unnecessary cognitive load. 
\textbf{\ding{183}} \textbf{Minimizing heterogeneity benefits complex domains.} The advantage of interface unification is most pronounced in demanding tasks like Financial Analysis (62.50\% vs. 47.50\%) and Repository Management (25.00\% vs. 14.19\%). By maintaining a single, expressive interface, the agent avoids action-space heterogeneity and can focus entirely on determining internal execution and task logic.
\begin{table}[t]
\centering
\small
\caption{Ablation study on execution interface design. \nosft{} denotes the instruction model without SFT, while \sft{} denotes SFT on the corresponding trajectory data.}
% \vspace{-0.3cm}
\label{tab:ablation_interface}
\resizebox{\columnwidth}{!}{
\begin{tabular}{c l c c c c c}
\toprule
\textbf{SFT} & \textbf{Interface} & \textbf{Fin.} & \textbf{Repo.} & \textbf{Loc.} & \textbf{Web.} & \textbf{Avg.} \\
\midrule
\nosft & ReAct & 32.50 & 7.14 & 25.71 & 16.00 & 20.92 \\
\sft & Atomic+HyperTool & 47.50 & 14.19 & 25.71 & 20.00 & 26.85 \\
\sft & HyperTool-only & 62.50 & 25.00 & 28.57 & 18.00 & 33.33 \\
\bottomrule
\end{tabular}
}
% \vspace{-0.2cm}
\end{table}

\begin{figure}[t]
    \centering
    % 请将 {example-image} 替换为你的真实图片路径，例如 {figures/token_analysis.pdf}
    \includegraphics[width=1\linewidth]{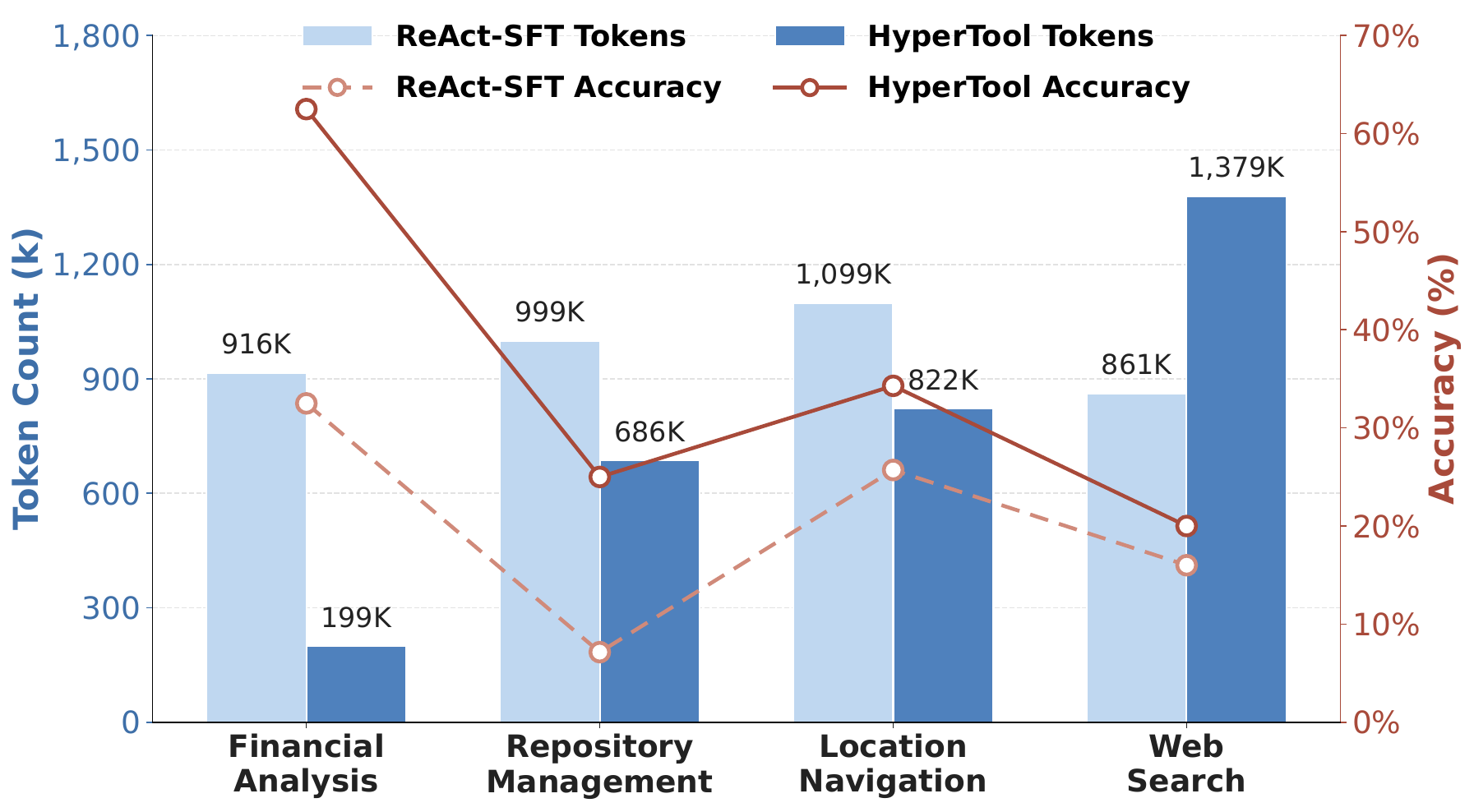}
    % \vspace{-0.8cm}
    \caption{Task Accuracy (\%) and Average Token Consumption (in thousands, k) comparison between ReAct-SFT and HyperTool-SFT on MCP Universe.}
    % \vspace{-0.3cm}
    \label{fig:token_analysis}
\end{figure}

\begin{figure*}[t]
\centering 
\includegraphics[width=1.\linewidth]{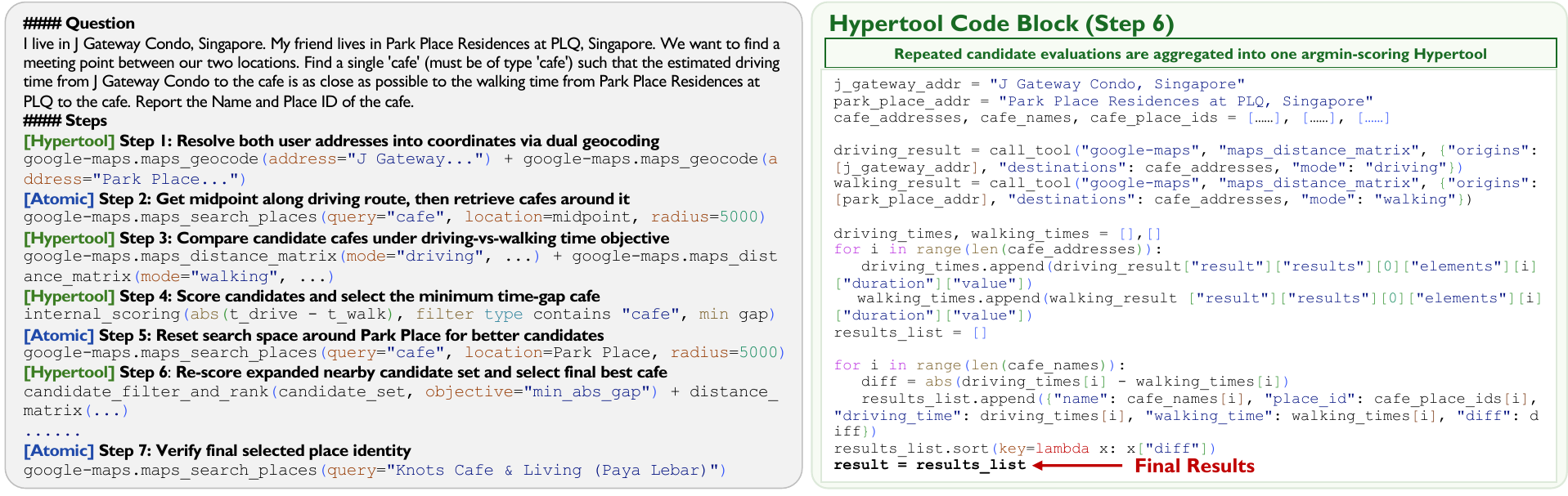} 
% \vspace{-0.8cm}
\caption{Case study of HyperTool on a location-navigation task.
The left panel shows the agent's reasoning trajectory, where standard atomic tool calls are interleaved with HyperTool blocks.
The right panel details the HyperTool code block generated at Step 6.
HyperTool locally batches, computes, and sorts tool outputs, returning only the task-relevant result while keeping intermediate procedural details out of the main trace.}
% \vspace{-0.3cm}
\label{fig:case_study} 
\end{figure*}

\noindent \textbf{Component Ablation.}
To evaluate the contribution of individual modules, we isolate the effects of \textit{Context Compression} and \textit{Local Repair} during the trajectory rollout stage, as well as \textit{Execution Filtering} and \textit{Evidence Filtering} during the trajectory verification stage. Figure~\ref{fig:ablation_study} presents the synthesis pass rates and downstream task accuracy under these configurations, revealing two critical insights: 
\textbf{\ding{182}} \textbf{Rollout modules prevent generation failure.} Removing \textit{Context Compression} severely drops the pass rate in context-heavy tasks (e.g., Web Search 70.0\%$\rightarrow$41.0\%) as the agent struggles with severe context inflation. Furthermore, removing \textit{Local Repair} leaves minor syntax errors unresolved; disabling both halves the success rate in Financial Analysis (40.0\%$\rightarrow$20.5\%). 
\textbf{\ding{183}} \textbf{Strict filtering guarantees downstream accuracy.} Omitting \textit{Execution} or \textit{Evidence Filtering} pollutes the SFT dataset with malformed or reasoning-inconsistent trajectories. This noise crashes the final average accuracy from 33.33\% down to 18.06\% and 21.05\%, respectively, demonstrating that rigorous verification is essential for effective model fine-tuning.

% \vspace{-0.1cm}
\subsection{Analysis}
\label{sec:analysis}

\paragraph{Token Efficiency.}
Figure \ref{fig:token_analysis} evaluates token consumption and task accuracy across all 253 evaluated trajectories from the MCP-Universe benchmark, comparing the ReAct-SFT baseline with HyperTool. The results reveal two clear trends: \textbf{\ding{182}} \textbf{HyperTool significantly improves overall efficiency.} It delivers higher average accuracy (33.33\% vs. 20.92\%) while reducing the global average token load (816k vs. 955k). This advantage is most striking in \textit{Financial Analysis}, where HyperTool nearly doubles ReAct's accuracy (62.50\% vs. 32.5\%) while using over 78\% fewer tokens (199k vs. 916k), validating that local execution boundaries effectively mitigate context inflation. \textbf{\ding{183}} \textbf{Increased token usage reflects deeper exploration.} Although HyperTool consumes more tokens in \textit{Web Search} (1,379k vs. 861k), this increase is not a regression. Instead, it reflects the agent successfully navigating deeper into complex multi-step workflows.

\begin{figure}[t]
    \centering
    \includegraphics[width=\linewidth]{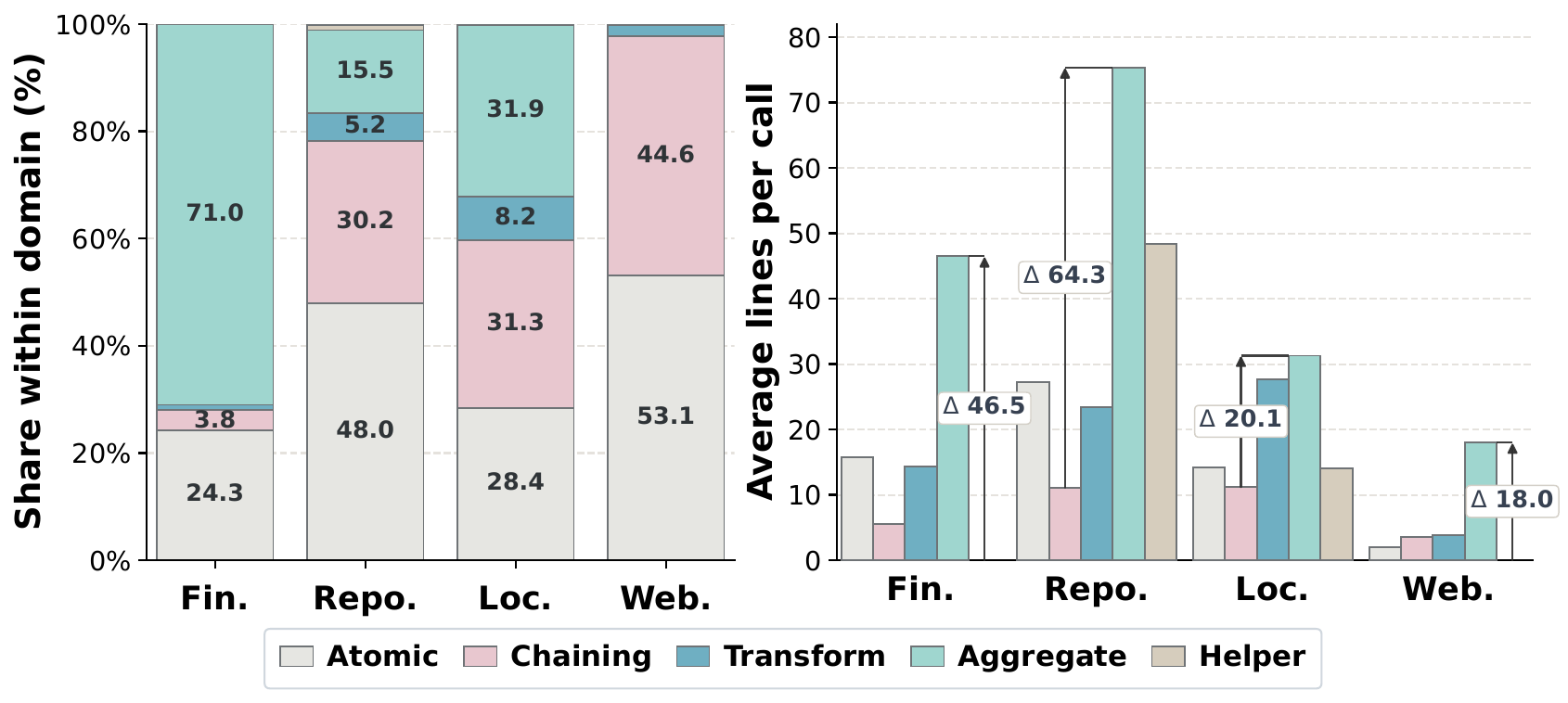}
    % \vspace{-0.8cm}
    \caption{Analysis of HyperTool's functional expansion and programmatic complexity.}
    % \vspace{-0.3cm}
    \label{fig:tool_expansion}
\end{figure}

\noindent \textbf{Tool-Behavior Analysis.}
To systematically analyze agent behavior, we categorize HyperTool blocks into five functional types—\textit{Atomic}, \textit{Chaining}, \textit{Transform}, \textit{Aggregate}, and \textit{Helper}—based on their operational roles (detailed descriptions are provided in Appendix~\ref{sec:block_types}). Figure~\ref{fig:tool_expansion} illustrates the usage distribution of these categories across different domains alongside their programmatic complexity (lines of code), revealing two major trends: 
\textbf{\ding{182}} \textbf{Orchestration of primitive tools.} Over half (55.7\%) of the blocks expand beyond a single \textit{Atomic} call, led by tool \textit{Chaining} (35.4\%). This confirms the model tends to synthesize workflows rather than relying on traditional atomic calls. 
\textbf{\ding{183}} \textbf{Internalization of procedural logic.} \textit{Aggregate} and \textit{Helper} blocks exhibit the highest complexity, averaging 44.73 and 42.67 lines of code (LOC). Across domains, \textit{Aggregate} blocks require more extensive implementations because the agent writes custom data processing algorithms to manipulate the results of tool calls, peaking in Repository Management at 75.4 LOC. This density demonstrates that intricate logic is fully digested locally, shielding the main trace from context inflation.

\noindent\textbf{Case Study.} To directly illustrate dynamic context management, Figure \ref{fig:case_study} captures the agent interleaving execution modes in a complex location task. For exploratory and verification actions (Steps 5 and 7), the agent uses standard \textbf{Atomic} calls to maintain high-level strategic control. However, for the intensive candidate evaluation (Step 6), it deploys a \textbf{HyperTool} block. This self-contained subroutine locally batches requests to the distance matrix MCP tool, computes absolute time gaps ($|t_{drive} - t_{walk}|$), and sorts candidates. By internalizing this dense procedural loop, the agent distills the optimal destination (``Knots Cafe \& Living'') while pruning verbose intermediate metrics. This powerfully demonstrates HyperTool's ability to bridge high-level planning with deterministic execution, rigorously shielding the main trace from context inflation.

% \vspace{-0.05cm}
\section{Conclusion}

We introduced \textbf{HyperTool}, a unified executable MCP-style tool interface that changes the model-visible unit of tool execution. By folding locally deterministic tool workflows into executable blocks, HyperTool keeps transient observations and value transfers inside the block while preserving high-level reasoning in the main trace. We construct verified HyperTool-format trajectories to teach models when and how to compose tools internally. Experiments on MCP-Universe show consistent gains over step-wise baselines, suggesting that tool-augmented agents should learn not only which tools to call, but also the appropriate granularity at which tool execution should be exposed.

% \newpage
\section*{Limitations}

HyperTool introduces a promising execution abstraction for tool-augmented agents by replacing stepwise atomic tool calls with executable blocks. However, its current design is most suitable for locally deterministic tool subroutines, such as retrieval, filtering, computation, and aggregation, where intermediate results can be processed within a block without frequent model-level intervention. 
% More interactive settings, such as real-time environments, tools with highly dynamic states, or workflows that require frequent human-like judgment, may require additional mechanisms for deciding when to expose intermediate observations back to the outer trace.
In addition, our training data covers a range of compositional tool-use tasks, but the diversity and scale of HyperTool trajectories can be further expanded. Larger trajectory collections across more domains, tool ecosystems, and task structures may improve the model's generalization to unseen workflows and help it learn more robust block-boundary decisions. We leave these extensions to future work.

% \section*{Acknowledgments}
% [Acknowledgments placeholder]

\bibliography{custom}

\appendix
\clearpage
\section{Experimental setup}
\subsection{SFT Setup}
\begin{figure}[htbp]
    \centering
    \includegraphics[width=1\linewidth]{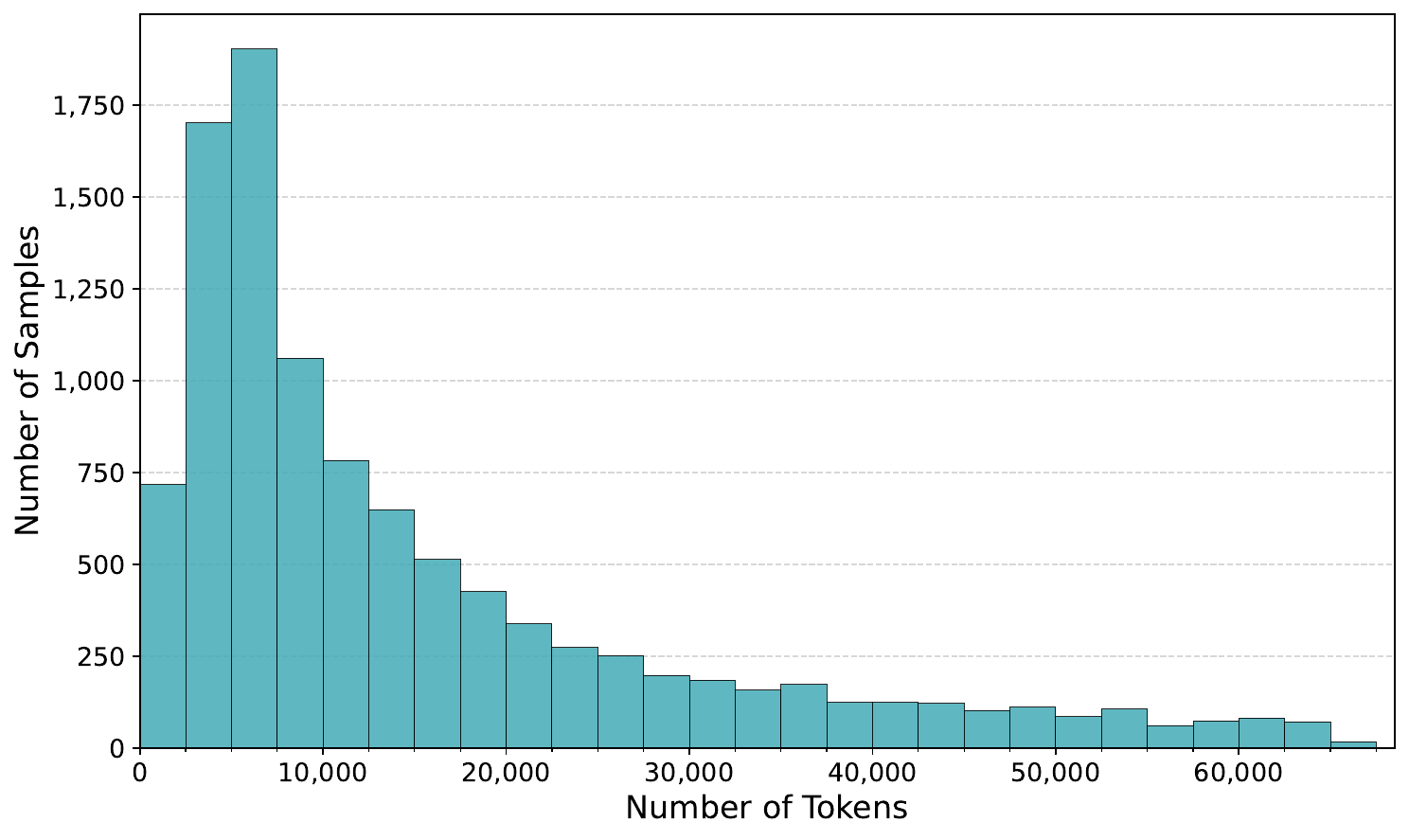}
    \caption{Token length distribution of the verified HyperTool SFT dataset.}
    \label{fig:token_dist}
\end{figure}

To construct a high-quality training set, we employ GLM-5.1 as the teacher model to synthesize compositional tasks and roll out trajectories, followed by GPT-4o acting as the trajectory judge to strictly verify execution and reasoning consistency. The final supervised fine-tuning (SFT) dataset comprises 10,422 verified HyperTool trajectories. As illustrated in Figure~\ref{fig:token_dist}, the token length of these trajectories exhibits a prominent long-tail distribution. While the majority of samples are concentrated between 5,000 and 15,000 tokens, a significant portion extends up to 60,000 tokens, reflecting the extensive context required for compositional multi-tool workflows.
% We fine-tune the Qwen3-8B and Qwen3-32B base models using a computing cluster equipped with 16 NVIDIA H200 GPUs.
To accommodate the long-tail token distribution of our dataset and efficiently manage the memory footprint of long-context training, our infrastructure is orchestrated via Ray and Megatron-LM. During training, the maximum sequence length is explicitly set to 65,536 tokens (64k context window).

For the hyperparameters, we use a global batch size of 64 and train the models for 3 epochs. The models are optimized using the Adam optimizer with a weight decay of 0.1 ($\beta_1=0.9$, $\beta_2=0.95$). The peak learning rate is set to 5e-5, following a cosine learning rate decay schedule with a 10\% warmup fraction and a minimum learning rate of 1e-6. Under this hardware and configuration, the total training time is approximately 4 hours for the Qwen3-8B model and 31 hours for the Qwen3-32B model.

\subsection{Evaluation Protocol}
We evaluate our models on the MCP-Universe benchmark under a strict zero-shot setting. For all baseline comparisons, we utilize the official default system prompts provided by the benchmark without any task-specific few-shot demonstrations. During inference, the generation hyperparameters are uniformly set to a temperature of 1.0, relying on the models' default top-p and top-k sampling configurations, with a maximum generation limit of 8,192 new tokens per turn.
To ensure standardized environment execution, we do not impose a global timeout on the overall task duration. However, strict execution boundaries are enforced at the tool interaction level: individual standard MCP tool calls are subject to a 120-second timeout, whereas HyperTool code block executions are allocated a timeout of up to 600 minutes to fully accommodate extensive local processing and data aggregation. 
For the final evaluation, the agent's ultimate responses are rigorously extracted from the generated <answer></answer> tags using regular expressions. Both Accuracy and Average Score metrics are subsequently computed using the official built-in evaluator provided by the MCP-Universe framework, ensuring a completely fair, objective, and reproducible assessment.

\section{HyperTool Interface}

\subsection{Taxonomy of HyperTool Blocks}
\label{sec:block_types}
To systematically analyze how the agent utilizes local execution boundaries, we classify the generated HyperTool code blocks into five functional categories based on their programmatic structure, tool orchestration complexity, and dataflow logic:
\textbf{Atomic.} The block serves as a minimal wrapper around a single primitive tool call without any additional logic or data manipulation. 

\textbf{Chaining.} The block executes a linear sequence of dependent tool calls, where the output of one tool is directly passed as an argument to a subsequent tool. By internalizing these linear dependencies, \textit{Chaining} blocks effectively hide intermediate procedural states from the main reasoning trace.

\textbf{Transform.} The block invokes one or more tools and subsequently applies deterministic data manipulation logic—such as parsing nested JSONs, filtering list elements, or extracting specific fields—before returning the result. It acts as a local data processor to prune verbose or noisy observations.

\textbf{Aggregate.} The block executes multiple tool calls (often via loops) and applies custom computational algorithms to aggregate their results. Typical operations include sorting candidates, finding minimum/maximum values , or computing derived metrics across multiple sources.

\textbf{Helper.} Representing the highest level of execution abstraction, the agent defines internal, reusable Python helper functions (`def helper():...`) within the block. This structure is deployed to manage highly repetitive operations, complex state transitions, or recursive data fetching across multiple primitive tools before yielding the final, distilled answer to the global context.

\subsection{HyperTool Examples}
\label{sec:hypertool_examples}
To provide a concrete illustration of how these blocks operate in practice and reduce the reasoning overhead, Figures~\ref{fig:case1_stepwise}--\ref{fig:case3_multicall} present a detailed comparison between a standard step-wise MCP trajectory and our internalized HyperTool execution blocks (including both atomic and multi-call scenarios).

\section{Additional Results and Case Studies}
\begin{figure}[htbp]
    \centering
    % 请将 {example-image} 替换为你的真实图片路径，例如 {figures/token_analysis.pdf}
    \includegraphics[width=1\linewidth]{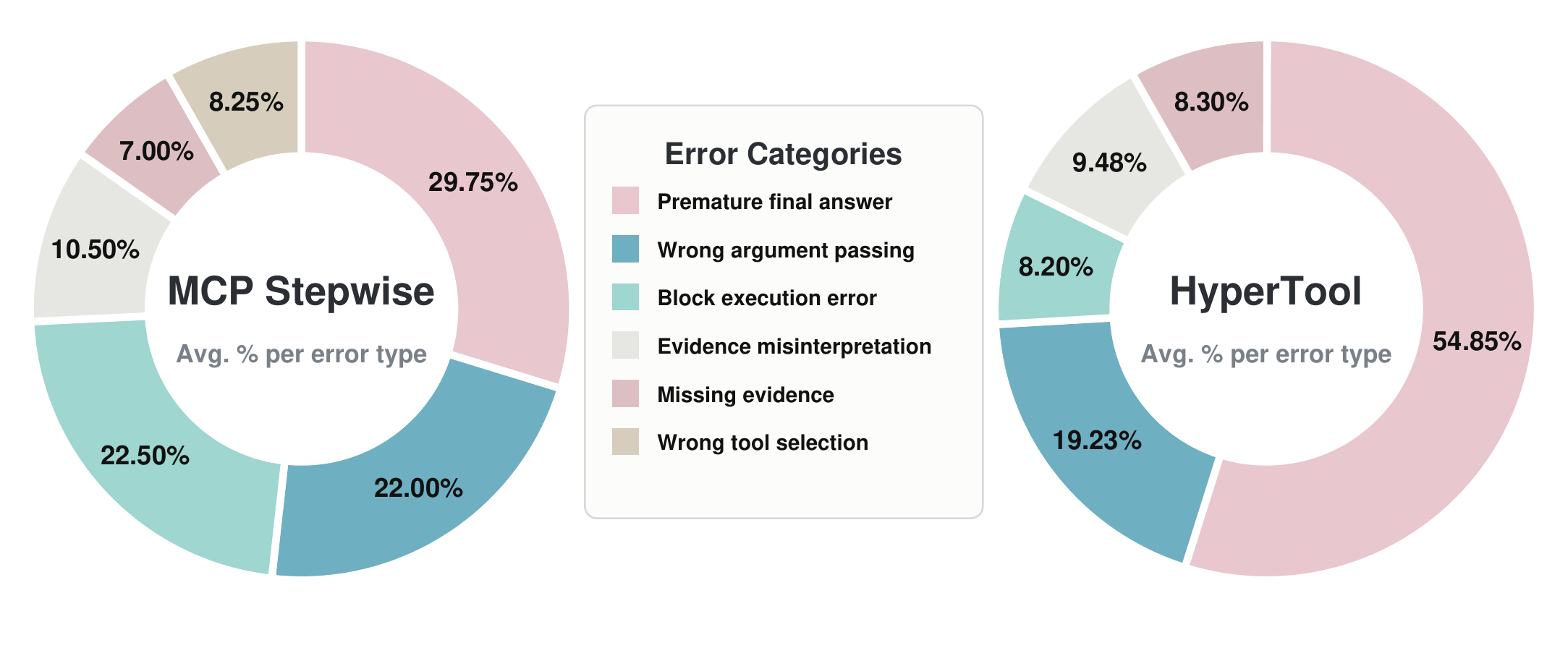}
    \caption{Distribution of failure modes for the MCP Stepwise baseline and HyperTool.}
    \label{fig:error_pie_charts}
\end{figure}

\paragraph{Error Analysis.} Figure \ref{fig:error_pie_charts} compares the error distribution between the MCP Stepwise baseline and HyperTool across all evaluated trajectories. The results demonstrate a fundamental shift in failure modes when utilizing local execution boundaries, characterized by three key observations: 
\textbf{\ding{182}} \textbf{Elimination of routing errors.} HyperTool completely eradicates \textit{Wrong tool selection} errors (8.3\%$\rightarrow$0.0\%). By encapsulating dependent operations within blocks, the model effectively bypasses intermediate routing confusion. 
\textbf{\ding{183}} \textbf{Significant reduction in execution crashes.} \textit{Block execution errors} drop substantially from 22.5\% to 8.2\%, confirming that delegating complex workflows to locally executed code reduces the cognitive burden of state tracking. 
\textbf{\ding{184}} \textbf{Shift towards premature termination.} As low-level mechanical errors are resolved, \textit{Premature final answer} emerges as the dominant failure mode, surging from 29.8\% to 54.9\%. This indicates that highly compressed, multi-step results can occasionally induce overconfidence, leading the agent to skip necessary secondary verifications. Furthermore, \textit{Wrong argument passing} remains relatively stable (22.0\%$\rightarrow$19.2\%), highlighting schema alignment as a persistent challenge for LLMs regardless of the execution paradigm.

\paragraph{Execution Dynamics Analysis.} Table~\ref{tab:tool_usage_stats} compares tool usage statistics across different methods based on Qwen3-8B backbone by measuring actual primitive tools executed (\# Tool using), model-visible actions (\# Tool calling), and total interaction turns. The granular breakdown reveals two clear trends: \textbf{\ding{182}} \textbf{Stepwise paradigms suffer from strict action coupling.} In standard stepwise baselines like ReAct-SFT, actual tool usage strictly mirrors model-visible calls (e.g., exactly 26.92 for both). This rigid one-to-one mapping forces every single tool operation to consume an interaction turn, inevitably leading to rapid context inflation. \textbf{\ding{183}} \textbf{HyperTool decouples task capacity from context overhead.} By internalizing dense tool workflows into local executable blocks, HyperTool orchestrates nearly double the number of primitive tools for deeper task exploration (26.92$\rightarrow$47.55). Despite this massive increase in actual tool utilization, it significantly reduces both external model-visible calls (26.92$\rightarrow$20.76) and overall interaction turns (28.81$\rightarrow$21.76). This empirical evidence directly confirms that HyperTool drastically expands the agent's operational capacity without inflating the global reasoning trace.

\begin{table}[t]
\centering
\small
\setlength{\tabcolsep}{3pt}
\begin{tabular}{llccc}
\toprule
Model & Method & \#Tool Using & \#Tool Calling & Turns \\
\midrule
\nosft & Base        & 5.58  & 5.58  & 6.84  \\
\nosft & AgentFold   & 5.38  & 5.38  & 11.78 \\
\nosft & Recode      & 12.72 & 12.72 & 16.20 \\
\nosft & Browsemaster & 5.84 & 5.84  & 7.85  \\
\nosft  & CodeAct     & 4.26  & 4.26  & 9.22  \\
\sft      & React       & 26.92 & 26.92 & 28.81 \\
\sft      & Hypertool   & 47.55 & 20.76 & 21.76 \\
\bottomrule
\end{tabular}
\caption{Comparison of tool usage statistics across base and SFT models. \nosft{} denotes the instruction model without SFT, while \sft{} denotes SFT on the corresponding trajectory data.}
\label{tab:tool_usage_stats}
\end{table}

\section{Potential Risk}
While HyperTool significantly enhances tool orchestration efficiency, it introduces specific challenges that warrant consideration. 
\textbf{1) Execution Security:} The core mechanism of HyperTool relies on the agent dynamically generating and executing Python code blocks. Without strict isolation, this exposes the host environment to Remote Code Execution (RCE) vulnerabilities, whether from malicious prompt injections or unintended infinite loops. Deploying HyperTool in real-world systems strictly necessitates robust sandboxing (e.g., Dockerized environments or restricted interpreters) and read-only API access constraints. 
\textbf{2) Interpretability Trade-offs:} By internalizing complex procedural logic into local blocks, HyperTool inherently hides intermediate execution states from the main reasoning trace. While this prevents context inflation, it makes diagnosing subtle algorithmic errors (e.g., an incorrect sorting condition within an \textit{Aggregate} block) more challenging, as the agent only observes the final output. 
\textbf{3) Reliance on Code Generation Capability:} The efficacy of HyperTool is bottlenecked by the base model's inherent coding proficiency. While our SFT pipeline effectively activates this capability in smaller models like Qwen3-8B, deploying this framework to zero-shot, non-code-optimized models may result in frequent syntax errors or schema hallucinations, necessitating robust local repair modules.

\section{Prompts}
In this section, we present prompts used in HyperTool , including HyperTool Agent which is used to define execution strategies and rules (\ref{sub:hyperTool}) , Trajectory Judge which is used to verify the reasoning chain and final answer (\ref{sub:Judge}) , HyperTool Code Repairer (\ref{sub:code_repair}) , Code Intent Reviewer (\ref{sub:intent_review}) , and Tool Response Summarizer (\ref{sub:summzrizer}).
\onecolumn
% ==================== CASE 1 (Figure) ====================
\begin{figure*}[tbp]
\centering
\begin{tcolorbox}[
    colback=white!95!gray,
    colframe=black,
    width=0.98\textwidth,
    arc=4mm,
    boxrule=0.5mm
]
\raggedright
\textbf{\large Case 1: Standard Step-wise MCP Trajectory}

% \vspace{0.5em}

\begin{lstlisting}[
    basicstyle=\scriptsize\ttfamily,
    breaklines=true,
    breakatwhitespace=true,
    columns=fullflexible,
    keepspaces=true,
    frame=single
]
Task: Find a single 'cafe' such that driving time from J Gateway Condo to the cafe is closest to walking time from Park Place Residences at PLQ to the cafe.

Step-wise MCP trajectory:
1. maps_geocode(address="J Gateway Condo, Singapore")
   -> lat=1.3357764, lng=103.7424381

2. maps_geocode(address="Park Place Residences at PLQ, Singapore")
   -> lat=1.3163901, lng=103.8927333

3. maps_search_places(query="cafe", location={"latitude": 1.32608325, "longitude": 103.8175857}, radius=5000)
   -> initial cafe candidates

4. maps_distance_matrix(origins=["J Gateway Condo, Singapore"], destinations=[initial cafe candidates], mode="driving")

5. maps_distance_matrix(origins=["Park Place Residences at PLQ, Singapore"], destinations=[initial cafe candidates], mode="walking")

6. Compare |driving_time - walking_time| for all initial candidates
   -> best initial candidate: Ms Durian | Speciality Bakery & Cafe, diff=1953

7. maps_search_places(query="cafe", location={"latitude": 1.3163901, "longitude": 103.8927333}, radius=5000)
   -> expanded candidate set near Park Place Residences at PLQ

8. maps_distance_matrix(origins=["J Gateway Condo, Singapore"], destinations=[expanded cafe candidates], mode="driving")

9. maps_distance_matrix(origins=["Park Place Residences at PLQ, Singapore"], destinations=[expanded cafe candidates], mode="walking")

10. Compare |driving_time - walking_time| for all expanded candidates
    -> best candidate: Knots Cafe & Living (Paya Lebar), driving_time=1812, walking_time=1654, diff=158

11. maps_search_places(query="Knots Cafe & Living (Paya Lebar)")
    -> types=["cafe", "restaurant", "point_of_interest", "store", "food", "establishment"]

Final answer:
{"stops": [{"name": "Knots Cafe & Living (Paya Lebar)", "place id": "ChIJKXkHO4sX2jERhWTgW68Gs44"}]}
\end{lstlisting}
\end{tcolorbox}
\caption{Trajectory trace for Case 1: Standard Step-wise MCP execution.}
\label{fig:case1_stepwise}
\end{figure*}

% ==================== CASE 2 (Figure) ====================
\begin{figure*}[tbp]
\centering
\begin{tcolorbox}[
    colback=white!95!gray,
    colframe=black,
    width=0.98\textwidth,
    arc=4mm,
    boxrule=0.5mm
]
\raggedright
\textbf{\large Case 2: One-call HyperTool Block}

% \vspace{0.5em}

\begin{lstlisting}[
    basicstyle=\scriptsize\ttfamily,
    breaklines=true,
    breakatwhitespace=true,
    columns=fullflexible,
    keepspaces=true,
    frame=single
]
Assistant reasoning: Let me start by geocoding both locations and searching for cafes.

HyperTool call:
j_gateway = call_tool("google-maps", "maps_geocode", {"address": "J Gateway Condo, Singapore"})
park_place = call_tool("google-maps", "maps_geocode", {"address": "Park Place Residences at PLQ, Singapore"})
result = {"j_gateway": j_gateway, "park_place": park_place}

Tool response:
{
  "j_gateway": {"status": true, "result": {"location": {"lat": 1.3357764, "lng": 103.7424381}, "formatted_address": "2 Gateway Dr, Singapore 608533", "place_id": "ChIJ3dTg3A8Q2jERI61lPg1ExlM"}},
  "park_place": {"status": true, "result": {"location": {"lat": 1.3163901, "lng": 103.8927333}, "formatted_address": "2 Paya Lebar Rd, Singapore 409053", "place_id": "ChIJ42A__8IZ2jERXCRt0_s38Nc"}}
}
\end{lstlisting}
\end{tcolorbox}
\caption{Trajectory trace for Case 2: One-call HyperTool Block (Atomic execution).}
\label{fig:case2_onecall}
\end{figure*}

% ==================== CASE 3 (Figure) ====================
\begin{figure*}[tbp]
\centering
\begin{tcolorbox}[
    colback=white!95!gray,
    colframe=black,
    width=0.98\textwidth,
    arc=4mm,
    boxrule=0.5mm
]
\raggedright
\textbf{\large Case 3: Multi-call HyperTool Block}

% \vspace{0.5em}

\begin{lstlisting}[
    basicstyle=\scriptsize\ttfamily,
    breaklines=true,
    breakatwhitespace=true,
    columns=fullflexible,
    keepspaces=true,
    frame=single
]
Assistant reasoning: Let me now calculate the driving time from J Gateway Condo and walking time from Park Place Residences for each of these cafes.

HyperTool call:
j_gateway_addr = "J Gateway Condo, Singapore"
park_place_addr = "Park Place Residences at PLQ, Singapore"

driving_result = call_tool("google-maps", "maps_distance_matrix", {"origins": [j_gateway_addr], "destinations": cafe_addresses, "mode": "driving"})
walking_result = call_tool("google-maps", "maps_distance_matrix", {"origins": [park_place_addr], "destinations": cafe_addresses, "mode": "walking"})

driving_times, walking_times = [], []
for i in range(len(cafe_addresses)):
    driving_times.append(driving_result["result"]["results"][0]["elements"][i]["duration"]["value"])
    walking_times.append(walking_result["result"]["results"][0]["elements"][i]["duration"]["value"])

results_list = []
for i in range(len(cafe_names)):
    diff = abs(driving_times[i] - walking_times[i])
    results_list.append({"name": cafe_names[i], "place_id": cafe_place_ids[i], "driving_time": driving_times[i], "walking_time": walking_times[i], "diff": diff})

results_list.sort(key=lambda x: x["diff"])
result = results_list

Tool response:
[
  {"name": "Knots Cafe & Living (Paya Lebar)", "place_id": "ChIJKXkHO4sX2jERhWTgW68Gs44", "driving_time": 1812, "walking_time": 1654, "diff": 158},
  {"name": "Keng Wah Sung Cafe", "place_id": "ChIJ47NO1xcY2jERyFbfhYX2QFU", "driving_time": 1669, "walking_time": 209, "diff": 1460},
  {"name": "The Upper Room Restaurant & Cafe", "place_id": "ChIJ0_8xYqAZ2jERpqkRsZovOZM", "driving_time": 1673, "walking_time": 3216, "diff": 1543}
]
\end{lstlisting}

\end{tcolorbox}
\caption{Trajectory trace for Case 3: Multi-call HyperTool Block.}
\label{fig:case3_multicall}
\end{figure*}
\subsection{Prompt Template of HyperTool Agent}\label{sub:hyperTool}

\begin{tcolorbox}[
    colback=white!95!gray, 
    colframe=black, 
    width=1\textwidth, 
    arc=4mm, 
    boxrule=0.5mm
]
\raggedright
You are a helpful assistant. Put your final answer in \texttt{<answer></answer>} part.

% \vspace{0.8em}
\textbf{\large HyperTool Strategy}

% \vspace{0.3em}
\textbf{IMPORTANT:} To maximize efficiency and reduce latency, you should prioritize HyperTool Strategy over sequential turn-based calls whenever the workflow is predictable.

% \vspace{0.8em}
\textbf{When to use HyperTool (Single code block)}
\begin{itemize}[leftmargin=1.5em, nosep]
    \item The task can be decomposed into a clear sequence of tool calls (A $\to$ B $\to$ C).
    \item The required data flow between steps is straightforward and can be handled with simple logic (e.g., filtering, extraction, loops).
    \item The structure of tool outputs is either:
    \begin{itemize}[nosep]
        \item Known from prior experience, OR
        \item Can be reliably inferred, OR
        \item Not critical to control flow (i.e., minor parsing errors will not break the pipeline).
    \end{itemize}
\end{itemize}

% \vspace{0.8em}
\textbf{When NOT to use HyperTool (Call tools one by one)}
\begin{itemize}[leftmargin=1.5em, nosep]
    \item The next step depends heavily on interpreting the semantics of previous outputs (non-trivial reasoning).
    \item The structure of tool outputs is unclear and critical for downstream steps.
    \item The workflow may require significant adaptation after observing intermediate results.
\end{itemize}

% \vspace{0.8em}
\textbf{Execution Standard}

% \vspace{0.3em}
\textbf{Important:}
\begin{itemize}[leftmargin=1.5em, nosep]
    \item If the logic is deterministic, do not wait for the system to respond between steps. Combine the calls, process the data locally within the block, and return the final consolidated result.
    \item Do not include any reasoning content in the code block, including as comments (e.g., lines starting with \texttt{\#}).
\end{itemize}

% \vspace{0.8em}
\textbf{Inference Key Rules:}
\begin{enumerate}[leftmargin=1.5em, nosep]
    \item The only tool you can call directly is \texttt{HyperTool}. Other tools can only be called through HyperTool by using \texttt{call\_tool(server\_name, tool\_name, \{"param": "value"\})}---note double braces in format string.
    \item You must carefully read the tool descriptions and the tool output format, and use the exact argument names as specified without deviation.
    \item Do not introduce unsupported facts; use reasoning to interpret tool results, but ensure conclusions are evidence-grounded. If a tool returns empty, \texttt{None}, or unexpected output, you MUST diagnose and retry with a corrected approach.
    \item The variables in different HyperTool code blocks are not reusable.
    \item Carefully follow the tool output format and implement simple data processing logic within the HyperTool code block to extract the required information. Only use fields explicitly provided in the tool output, and eliminate any irrelevant or noisy data.
    \item Remember each HyperTool code block must use \texttt{"result"} as the variable name to store the final result. \texttt{"result"} must contain the actual data returned by tool calls (or processed/extracted values derived from it), never a hand-written status string like \texttt{"done"} or \texttt{"completed"}. Do not use print statements to print the result.
\end{enumerate}
\end{tcolorbox}

\subsection{Prompt of Trajectory Judge}\label{sub:Judge}
\begin{tcolorbox}[
    colback=white!95!gray,
    colframe=black,
    width=1\textwidth,
    arc=4mm,
    boxrule=0.5mm,
    breakable
]
\raggedright
\textbf{\large Task}

% \vspace{0.3em}
\$QUESTION\$

% \vspace{0.8em}
\textbf{\large Trajectory}

% \vspace{0.3em}
\$TRAJECTORY\$

% \vspace{0.8em}
\textbf{\large Agent's Final Answer}

% \vspace{0.3em}
\$AGENT\_ANSWER\$

% \vspace{0.8em}
\textbf{\large Evaluation Steps}

% \vspace{0.5em}
\textbf{Step 1: Reconstruct the final reasoning chain}

% \vspace{0.3em}
Trace backward from the final answer to find the minimal set of tool calls, code executions, and reasoning steps that directly support it. Ignore earlier abandoned or corrected attempts.

% \vspace{0.5em}
\textbf{Step 2: Verify each step in the final chain}

% \vspace{0.3em}
For each step, check:
\begin{itemize}[leftmargin=1.5em, nosep]
    \item Is the Python code logically correct for its stated purpose?
    \item Does the agent correctly read and interpret the tool output?
    \item Are intermediate conclusions consistent with the tool outputs?
\end{itemize}

% \vspace{0.5em}
\textbf{Step 3: Check for selective reporting / post-hoc rationalization}

% \vspace{0.3em}
This is critical. Ask:
\begin{itemize}[leftmargin=1.5em, nosep]
    \item Did the agent compute multiple variants of a metric (e.g., different date ranges, open vs.\ close, close vs.\ intraday high) and then selectively report only the variant that supports its conclusion?
    \item Does the agent's chosen metric definition match the standard interpretation implied by the task wording? For example:
    \begin{itemize}[nosep]
        \item ``price increased from date A to date B'' $\to$ should use closing prices on those two specific dates, not open-to-intraday-high or any other cherry-picked combination.
        \item ``revenue in Q2'' $\to$ should use the Q2 figure, not an annualized estimate.
    \end{itemize}
    If the agent uses a non-standard definition without explicit justification, treat this as a failure even if the arithmetic is correct.
    \item If the agent tried multiple definitions and only one of them satisfies the task condition, but the agent presents that one as the answer without acknowledging the others failed, this is selective reporting $\to$ score 0.
\end{itemize}

% \vspace{0.5em}
\textbf{Step 4: Verify the final answer}
\begin{itemize}[leftmargin=1.5em, nosep]
    \item Does the final answer directly follow from the last valid reasoning step?
    \item Is the stated conclusion actually supported by the data, under the standard metric definition (not a cherry-picked one)?
\end{itemize}

% \vspace{0.5em}
\textbf{Step 5: Check completeness}
\begin{itemize}[leftmargin=1.5em, nosep]
    \item Does the final answer directly and completely address every sub-question in the task?
    \item Is the answer specific enough (concrete values, names, conclusions)?
\end{itemize}

% \vspace{0.8em}
\textbf{\large Scoring Rules}

% \vspace{0.5em}
Score \textbf{1} only if ALL of the following hold:
\begin{itemize}[leftmargin=1.5em, nosep]
    \item The final reasoning chain is logically valid end-to-end
    \item The agent correctly interprets all tool outputs in the chain
    \item The final answer is consistent with the chain's conclusion
    \item The metric / method / source used matches the standard interpretation implied by the task (no cherry-picking of favorable definitions)
    \item No evidence of selective reporting: the agent did not try multiple metric variants and report only the one that fits
    \item Any task-specified method or source is visibly followed
    \item The final answer is complete and directly addresses the task
\end{itemize}

% \vspace{0.5em}
Score \textbf{0} if ANY of the following:
\begin{itemize}[leftmargin=1.5em, nosep]
    \item A logical or computational error exists in the final reasoning chain
    \item The agent misreads a tool output (wrong key, wrong type, off-by-one, etc.)
    \item The final answer contradicts the reasoning chain
    \item A clearly stated required method / source was ignored or substituted
    \item The agent uses a non-standard metric definition without justification (e.g., open-to-intraday-high instead of close-to-close for a price change claim)
    \item The agent computed multiple metric variants and selectively reported only the one supporting its preferred conclusion
    \item The trajectory provides no evidence supporting the final answer
    \item The final answer is incomplete, vague, or missing required parts
\end{itemize}

% \vspace{0.5em}
\textbf{Note on truncated tool outputs:} Judge based on what is visible. Do not penalize for truncation unless the missing part was demonstrably necessary to reach the answer.

% \vspace{0.8em}
\textbf{\large Output Format}

% \vspace{0.3em}
Return exactly this JSON (no markdown, no extra text):
% \vspace{0.3em}

\texttt{\{} \\
\hspace*{1em}\texttt{``score'': 0 or 1,} \\
\hspace*{1em}\texttt{``selective\_reporting\_check'': ``One sentence: did the agent try multiple metric} \\
\hspace*{3em}\texttt{definitions? Which ones passed/failed? Was only the passing one reported?'',} \\
\hspace*{1em}\texttt{``metric\_definition\_check'': ``One sentence: does the agent's chosen metric match} \\
\hspace*{3em}\texttt{the standard interpretation of the task wording?'',} \\
\hspace*{1em}\texttt{``reason'': ``2--4 sentences explaining the key reason for the score.} \\
\hspace*{3em}\texttt{Cite the specific step or code that passed or failed.''} \\
\texttt{\}}

\end{tcolorbox}

\subsection{Prompt of HyperTool Code Repairer}\label{sub:code_repair}
\begin{tcolorbox}[
    colback=white!95!gray,
    colframe=black,
    width=1\textwidth,
    arc=4mm,
    boxrule=0.5mm,
    breakable
]
\raggedright
You are a code repair assistant. A HyperTool code block failed during execution. Your job is to fix the code so it achieves the same goal as described in the original reasoning. You may add necessary exploratory steps (e.g., listing a directory before fetching a file) if they help reach the same goal. Do NOT change the final goal or overall strategy.

% \vspace{0.8em}
\textbf{\large Available Tools}

% \vspace{0.3em}
\$TOOLS\_INFO\$

% \vspace{0.8em}
\textbf{\large Reasoning that led to this code}

% \vspace{0.3em}
\$THINKING\$

% \vspace{0.8em}
The code above produced an error.

% \vspace{0.3em}
\textbf{Error details:} \$ERROR\_CONTENT\$

% \vspace{0.5em}
Determine whether this error can be fixed while keeping the same overall goal as the original reasoning.

% \vspace{0.8em}
\textbf{Allowed fixes}
\begin{itemize}[leftmargin=1.5em, nosep]
    \item Fix argument names, data types, syntax errors, or parsing logic.
    \item Add exploratory intermediate steps WITHIN the same code block (e.g., list a directory to discover the correct file path, then fetch the file), as long as the code still completes the original final goal afterward.
\end{itemize}

% \vspace{0.5em}
\textbf{Not allowed}
\begin{itemize}[leftmargin=1.5em, nosep]
    \item Add reasoning content or comments inside the code block (e.g., lines starting with \texttt{\#}).
    \item Change the final goal (e.g., targeting a different resource, repo, or object).
    \item Abandon the original strategy and replace it with something unrelated.
\end{itemize}

% \vspace{0.8em}
\textbf{Your response}
\begin{itemize}[leftmargin=1.5em, nosep]
    \item If the error \textbf{can} be fixed $\to$ call \texttt{HyperTool\_\_HyperTool} with the corrected code.
    \item If the error \textbf{cannot} be fixed (broken tool / network / auth / permission / rate limit / service unavailable / tool does not exist) $\to$ do NOT call any tool, respond with exactly: \texttt{False}
\end{itemize}
\end{tcolorbox}

\subsection{Prompt of Code Intent Reviewer}\label{sub:intent_review}
\begin{tcolorbox}[
    colback=white!95!gray,
    colframe=black,
    width=1\textwidth,
    arc=4mm,
    boxrule=0.5mm,
    breakable
]
\raggedright
You are a code intent reviewer. A HyperTool code block failed and was rewritten by a repair assistant. Your job: judge whether the fixed code preserves the SAME final goal as described in the original reasoning.

% \vspace{0.8em}
\textbf{\large Original reasoning (the intent behind the original code)}

% \vspace{0.3em}
\$REASONING\$

% \vspace{0.8em}
\textbf{\large Original code (before fix)}

% \vspace{0.3em}
\$ORIG\_CODE\$

% \vspace{0.8em}
\textbf{\large Fixed code (after fix)}

% \vspace{0.3em}
\$NEW\_CODE\$

% \vspace{0.8em}
\textbf{\large Judgment Criteria}

% \vspace{0.5em}
\textbf{VALID} if:
\begin{itemize}[leftmargin=1.5em, nosep]
    \item The fixed code targets the same resource / API / object as the original.
    \item The final operation type (fetch / write / compute / query) is unchanged.
    \item Only argument values, syntax, variable names, or parsing logic changed.
    \item Exploratory steps were added (e.g., list directory first), but the original goal is still completed afterward.
\end{itemize}

% \vspace{0.5em}
\textbf{INVALID} if:
\begin{itemize}[leftmargin=1.5em, nosep]
    \item The target resource, repo, endpoint, or object changed.
    \item The final operation type changed (e.g., fetch $\to$ hardcode, write $\to$ read).
    \item The original goal was dropped or replaced with a different task.
    \item The code only does exploration and never completes the original goal.
\end{itemize}

% \vspace{0.8em}
Reply strictly in one of these two formats (no other text):

% \vspace{0.3em}
\texttt{VALID: <one sentence reason>}\\
\texttt{INVALID: <one sentence reason>}
\end{tcolorbox}

\subsection{Prompt of Tool Response Summarizer}\label{sub:summzrizer}
\begin{tcolorbox}[
    colback=white!95!gray,
    colframe=black,
    width=1\textwidth,
    arc=4mm,
    boxrule=0.5mm,
    breakable
]
\raggedright
You are a precise information extractor. Your task: Extract and retain ONLY the information from the tool response that is related to the tool's intent and potentially useful for solving the original query. Remove all irrelevant, redundant, or unhelpful content.

% \vspace{0.5em}
Follow these rules:
\begin{enumerate}[leftmargin=1.5em, nosep]
    \item \textbf{Reference the query and history:} Use the original query and historical interactions to understand what information might be useful for solving the problem.
    \item \textbf{Conservative retention:} When in doubt, preserve information instead of deleting it.
    \item \textbf{Make it informative but concise:} Keep useful URLs and key evidence. Remove formatting noise and irrelevant verbosity.
    \item \textbf{Preserve numerical precision:} All numbers (prices, coordinates, IDs, percentages, counts, etc.) must be copied verbatim from the source. Never round, abbreviate, or paraphrase numeric values.
    \item \textbf{Handle unhelpful responses:} If fully unhelpful, return a short statement like ``No helpful information found.''
\end{enumerate}

% \vspace{0.8em}
\textbf{\large Original Task}

% \vspace{0.3em}
\$QUESTION\$

% \vspace{0.8em}
\textbf{\large Conversation History (for context)}

% \vspace{0.3em}
\$HISTORY\$

% \vspace{0.8em}
\textbf{\large Tool Call That Produced This Response}

% \vspace{0.3em}
\textbf{Tool:} \$TOOL\_NAME\$ \\
\textbf{Arguments:} \$TOOL\_ARGS\$

% \vspace{0.8em}
\textbf{\large Raw Tool Response to Summarize}

% \vspace{0.3em}
\$TOOL\_RESPONSE\$

% \vspace{0.8em}
Please provide a concise summary of the tool response above, retaining all information relevant to the original task.
\end{tcolorbox}

%\section{Case Show}
%\label{sec:case_show}

%\lstinputlisting[
    %basicstyle=\tiny\ttfamily,
    %breaklines=true,
   % breakatwhitespace=true,
    %columns=fullflexible,
    %keepspaces=true,
    %frame=single,
    %caption={Complete trajectory for \texttt{google\_maps\_task\_0028.json}.},
    %label={lst:case_show_google_maps_0028}
%]{google_maps_task_0028.json}

\end{document}